\documentclass[letterpaper, 10 pt, conference]{ieeeconf}  

\IEEEoverridecommandlockouts                              

\usepackage{times} 
\usepackage{amsmath} 
\usepackage{amssymb}  

\usepackage{graphicx}
\usepackage{algorithm,algpseudocode}
\usepackage[caption=false]{subfig}
\usepackage{csquotes}
\usepackage{color}
\usepackage{multirow}

\newcommand{\R}{\mathbb{R}}

\title{\LARGE \bf
A Flexible and Robust Vision Trap for Automated Part Feeder Design}

\author{Rasmus Laurvig Haugaard, Thorbj{\o}rn Mosekj{\ae}r Iversen, Anders Glent Buch, \\ Aljaz Kramberger and Simon Faarvang Mathiesen
\thanks{
This work was supported by Innovation Fund Denmark through the project MADE FAST. \newline
All authors are from SDU Robotics, Maersk Mc-Kinney Moller Institute, University of Southern Denmark.\newline
{\tt\small \{rlha,thmi,anbu,alk,simat\}@mmmi.sdu.dk}}
}

\begin{document}
\maketitle
\thispagestyle{plain}
\pagestyle{plain}

\begin{abstract}
Fast, robust, and flexible part feeding is essential for enabling automation of low volume, high variance assembly tasks. 
An actuated vision-based solution on a traditional vibratory feeder, referred to here as a vision trap, should in principle be able to meet these demands for a wide range of parts. 
However, in practice, the flexibility of such a trap is limited as an expert is needed to both identify manageable tasks and to configure the vision system.
We propose a novel approach to vision trap design in which the identification of manageable tasks is automatic and the configuration of these tasks can be delegated to an automated feeder design system. 
We show that the trap's capabilities can be formalized in such a way that it integrates seamlessly into the ecosystem of automated feeder design.
Our results on six canonical parts show great promise for autonomous configuration of feeder systems.

\end{abstract}


\section{Introduction}

An essential part of any robotics-driven assembly system is the feeding of parts used in the assembly into the system. For practical reasons, it is not uncommon for parts to arrive in unsorted bins, which leaves the task of singulation and recovering of the part pose to the robotic system. 
Traditionally, the task has been solved using manually designed vibratory feeders
that utilize vibration for part conveying and mechanical orientation techniques (often referred to as traps)
to ensure parts can be picked from a known location in a certain orientation. However, these feeders are, due to the complexity of their design, expensive to develop and designed specifically for the part for which they are intended. For this reason, the robotic vision communities have dedicated considerable efforts toward the development of flexible vision-based bin-picking solutions. 
While recent years have shown great advancements in the robustness of such systems, they are still too unreliable and too slow for many industrial cases~\cite{hodavn2020bop}.

\begin{figure}
    \centering
    \includegraphics[width=0.8\linewidth, trim={150, 500, 0, 300}, clip]{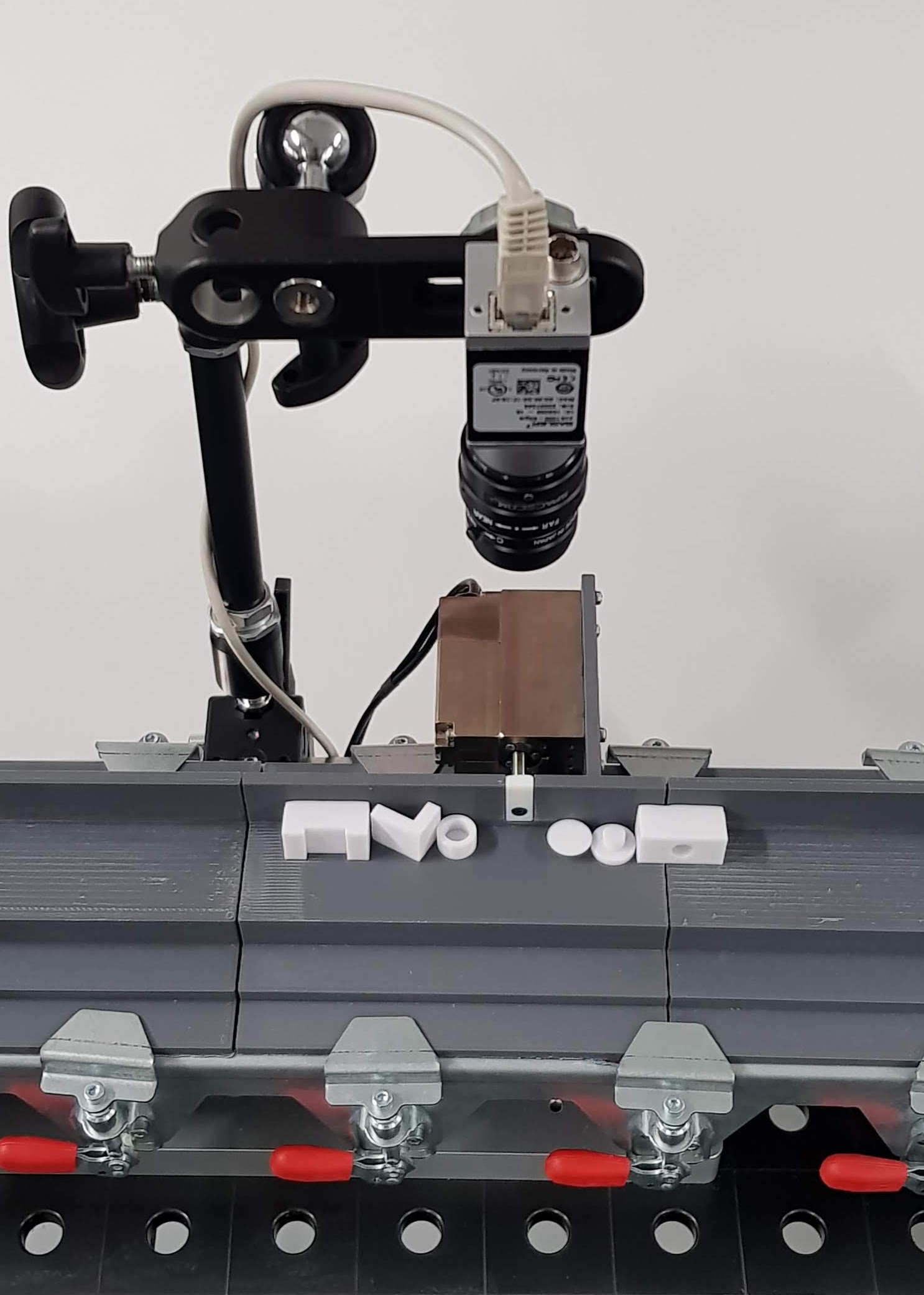}
    \caption{The proposed vision trap triggers a linear actuator to push parts off the vibratory feeder track unless the vision system determines that a part is oriented in a desired stable pose.
    Note that the image shows all the part types used in the experiments, but only instances of one part type is on the track at a time at run time.
    }
    \label{fig:visiontrap}
\end{figure}

\begin{figure*}
    \centering
    \includegraphics[width=.85\textwidth]{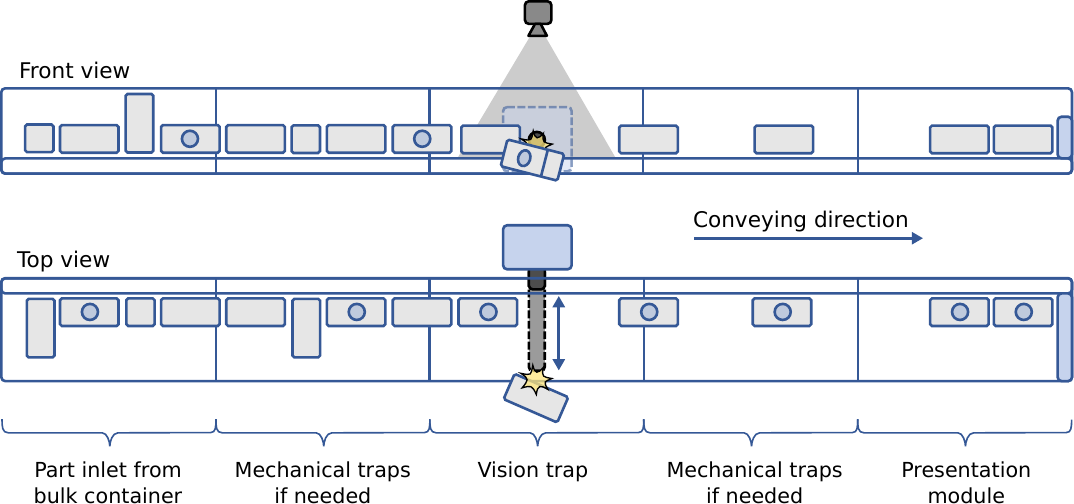}
    \caption{
    The vision trap in its proposed context.
    Parts enter the feeder from a bulk container and are conveyed along the tracks by vibration.
    Parts are rejected by the vision trap using a linear actuator, unless the vision system believes the part is in a desired orientation.
    Parts with desired orientations end up at the presentation module from which they can be picked by a robotic manipulator. 
    Mechanical traps can be added before and after the vision trap to reject or reorient parts beyond the capabilities of the vision trap. Recirculation of rejected parts is omitted for clarity.
    }
    \label{fig:visiontrapConcept}
\end{figure*}

Combining the flexibility of vision systems and the robustness of mechanical traps can to some extent be realized with vision traps. A vision trap is a special type of trap that consists of a vision sensor and a mechanical manipulation mechanism. The vision sensor detects if a part has a desired orientation and the manipulator interacts with the part based on the vision input. However, the development of vision traps is, to the best of our knowledge, still relying on manual configuration. 
First an expert is needed to manually identify which tasks are expected to be solvable by a vision system, after which the vision system is configured by the expert~\cite{hagelskjaer2018does}. 
This is not only an issue because manual configuration is costly,
in fact the reliance on manual configuration becomes an obstacle in the integration of vision traps into automated feeder design systems.

The foundation of this work is the realization that a vision trap is not an isolated entity that should be optimized on its own for a specific task. 
Instead, the vision trap should expose a configuration interface to the feeder design system, enabling the design optimization to combine the vision trap with mechanical traps in such a way that robustness is ensured. 
We demonstrate a novel approach to vision trap design which facilitates the fusion of vision traps with recent advances in automated vibratory feeder design~\cite{mathiesen2018optimisation}, which have shown that both the optimization of individual mechanical traps as well as the sequencing of these traps on the feeder can be automated~\cite{mathiesen2017automatic}. 
Our main contributions are as follows:

\begin{itemize}
    \item A novel vision trap that is automatically configurable for a variety of feeding tasks and has inherent estimates of its performance on said tasks.
    \item A formalization of the vision trap's capabilities that allows it to integrate seamlessly with the automated feeder design framework in~\cite{mathiesen2018optimisation}.
\end{itemize}

The proposed vision trap is realized through the use of dynamic simulation to infer the possible part orientations
and the use of convolutional neural networks, trained on synthetic data, to detect and classify the stable poses of parts in the trap. 
Inference on the synthetic images are then used to identify which stable poses the network is able to distinguish between, 
and, for a given configuration, this knowledge is formulated as a stochastic state transition matrix, 
which forms the basis of the simulation-based feeder design framework.

The setup is shown in Fig.~\ref{fig:visiontrap} and \ref{fig:visiontrapConcept}.
The performance of the vision trap is evaluated on 3D printed parts, where each part is representative of a canonical group of similar parts~\cite{Boothroyd2005}. 
We show that the automatically configured vision trap performs robustly and that it integrates well with automated feeder design.

\section{Related work}
Feeding of parts into a robotic assembly system is the process of singulating parts, recovering their orientation, and presenting the parts to a robotic manipulator. The ideal solution to part feeding would be fast, flexible, and reliable bin-picking, which is when parts can be introduced to the system unsorted in a bin and picked straight from that bin by a robotic manipulator guided by a vision sensor. However, flexible and reliable bin-picking is still an active research topic~\cite{hodavn2020bop}. To meet the industry's demand for flexible feeding, feeders have been developed which rely on vibrations to singulate parts on a surface such that the parts' poses can be estimated using a vision sensor. While this reduces the difficulty of the vision task, they still rely on an expert to identify which problem to solve and to configure the vision system. For a recent survey on flexible feeding, see~\cite{malik2019advances}. It is interesting to note that the academic community tends to focus on challenging benchmarks, where state of the art performance is still less than what is required for industrial application~\cite{hagelskjaer2019using}. This calls for more research into robust vision solutions. In this regard it has been recognized that quantification of uncertainties in a vision system is needed to ensure reliability~\cite{okorn2020learning}.

An alternative to the use of vision solutions for part feeding is the widely used vibratory feeders. There has been extensive work on formalizing the design of vibratory feeders. Most notable is the work~\cite{Boothroyd2005}, which in addition to an extensive analysis on vibratory feeder dynamics, provides a compendium on mechanical orienting devices together with guidelines for adjustment of these based on part geometry and operational parameters. In earlier work~\cite{murch1971predicting} they were the first to show that trap behavior can be described using stochastic state transition matrices,
encapsulating how a trap will reorient parts from one stable pose/orientation to another with a certain probability, or alternatively cause the part to fall off the feeder track on which it travels. The latter is also referred to as a rejection type trap. 
The transition matrices of a sequence of traps can then be chained to describe their combined behaviour.

Both~\cite{christiansen1996automated} and~\cite{mathiesen2017automatic} show how the matrices can be part of an automatic design algorithm for feeders and~\cite{mathiesen2017automatic} additionally shows how the required data can be obtained from dynamic simulation. In~\cite{mathiesen2018optimisation} the idea of simulation-based design is taken a step further with an efficient method for optimization of traps to achieve a target trap behavior and subsequently generate its transition matrix. 
However, for reducing the solution space to a manageable size some expert knowledge is needed to choose which target behaviors are worth optimizing towards. The strength of the specialized mechanical traps is also their weakness in terms of adding design complexity. This is due to the fact that, in order to achieve a working design, both the combinatorial problem of trap sequencing and trap behavior optimization must be handled holistically. An effort could be made to incorporate the design heuristics possessed by the human designers, however, an automated approach to reduce the solution space could make an exhaustive approach to automated design computationally viable.

Both~\cite{edmondson2001flexible} and~\cite{loy2010new} discuss approaches to how sensor-based solutions can help to simplify part feeder design. Additionally, while simple vision-based solutions are a staple in the industry, there exists recent work~\cite{kolditz2021flexible} on elaborate vision traps using pressurized air to actively reorient parts according to input from the vision system. The scope of the application for that work is outlined in~\cite{kolditz2021flexible}, however, is yet to be evaluated thoroughly.

Similarly to~\cite{loy2010new} we propose that a simple vision-based trap with a rejection mechanism, together with a few mechanical traps, can cover a significant portion of the feeding requirements found in the industry. In contrast to \cite{loy2010new}, our method is configured based on simulation and thus does not require collection or annotation of real images. 
Importantly, this has the consequence that while \cite{loy2010new} will typically be configured to a manually chosen trap behaviour, our method can automatically provide a large set of possible trap behaviours, in the form of transition matrices, that can integrate seamlessly with a system for automated feeder design.


There are many existing works on pose estimation \cite{posecnn, surfemb, sundermeyer}, but they are designed to work on the full, continuous rotation space, $SO(3)$. 
In this case however, we are only interested in estimating a discrete distribution over relatively few stable poses, which is a much simpler problem, allowing us to explicitly train a model to estimate said distribution.

\section{Method}
\label{sec:method}
The feeder design framework in~\cite{mathiesen2018optimisation}, which the vision trap has been developed to integrate with, is concerned with the changes in part orientations as parts pass through a trap. The physically permissible poses are denoted stable poses and the transitions between stable poses are formalized through the use of transition matrices. 
This section first presents an overview of the vision trap.
Then stable poses are defined and identified as used in this work,
followed by a description of how vision models are trained to detect and classify the stable poses.
This is followed by a short summary of the automated feeder design system,
and how the vision trap integrates with it by defining the vision trap transition matrix.
Lastly, as objects may contain many stable poses, making the subsequent optimization by the feeder designer infeasible, a method to reduce the number of configurations based on visual ambiguity is also presented.

\subsection{Overview}
Our vision trap is visualized in Fig.~\ref{fig:visiontrap} and \ref{fig:visiontrapConcept}. It consists of a feeding section, which is a 3D printed track with a flat surface slightly angled such that parts come to rest against the wall during vibration. A camera is mounted above the feeding section, and a linear actuator is mounted in the wall perpendicular to the feeding direction such that it can discard objects in undesirable stable poses. The vision pipeline consists of two deep learning based models: a detector and a stable pose classifier. When an instance is detected in front of the actuator, the stable pose classifier is used to estimate how probable it is that the given instance is in a desired stable pose. If this probability is not sufficiently high, the actuator is triggered which pushes the part off the track.


\subsection{Stable pose}
During operation, parts on the vibratory feeder will move along the feeding section. The combined normal forces from the surface and the wall constrain the possible orientations the parts. To identify these orientations, we use dynamic simulation of parts on a vibrating feeding section given a random pose initialization.
Multiple part instances are simulated simultaneously to encapsulate the interactions between parts.
As the simulation progresses, the instances converge to specific regions in $SO(3)$. Each such region is considered a single stable pose.

Specifically, after the objects have had time to settle, we discretize $SO(3)$, let the occurring poses be represented by their nearest neighbour in the discrete set, and remove all poses from the discrete set that have not occurred.
In an undirected graph, where each of the occurring discrete poses is a node, two poses are connected with an edge if the angle between them is below some threshold, and the stable poses are then defined as the connected components in the graph.
We denote the set of stable poses $S = \{s_1, s_2, ..., s_n\}$, where each $s_i$ is a connected component in the graph, and the prior over stable poses $P(s_i)$ is estimated from the occurrences in the dynamic simulation.

\subsection{Vision models and training details}
We train a detector, Faster R-CNN~\cite{faster-rcnn}, and a stable pose classifier, a ResNet50~\cite{resnet}, on synthetic data.
The classifier is fed with crops from the detector resized to 224x224 pixels and estimates a discrete probability distribution across the stable poses, conditioned on the image crop, $P(s_i| I)$.
The backbones of both models are pre-trained on ImageNet~\cite{imagenet} classification.

\begin{figure}
    \centering
    \includegraphics[width=0.8\linewidth, trim={180, 170, 0, 120}, clip]{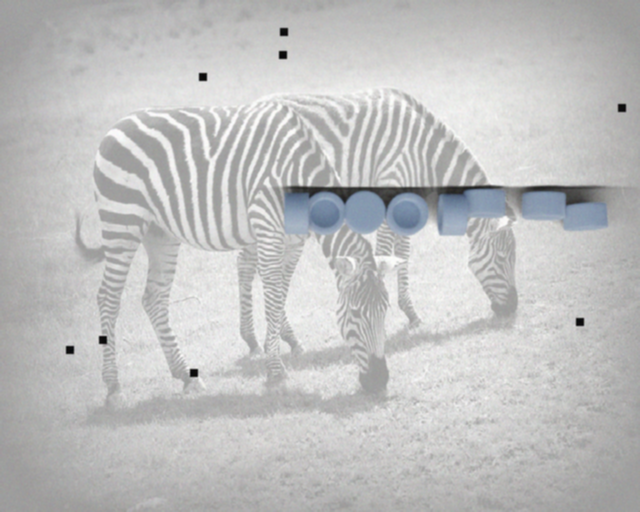}
    \caption{
        Example of synthetic training image with augmentations. 
        The target detections and stable pose classes comes automatically from the simulation.
    }
    \label{fig:synth_example}
\end{figure}

We use Blender\footnote{https://www.blender.org} to synthesize images based on dynamic simulation.
To reduce the reality gap, we employ domain randomization by sampling environment maps from freely available\footnote{https://polyhaven.com/hdris} HDRI maps to provide variety in lightning, and randomizing the material roughness and specularity.
Live during training, we sample background images from the COCO dataset \cite{coco-dataset} and add common augmentations including Gaussian Noise, Random Blur and Color Jitter.
A synthetic, augmented image example is shown in Fig.~\ref{fig:synth_example}.

In our experiments, we render 1000 synthetic images per object with multiple instances per image.
Both models are trained on the synthetic data using Adam~\cite{adam} with a learning rate of $10^{-4}$.
The detection and classification models are trained with a batch size of 4 and 8, for 20 and 50 epochs, respectively.

\subsection{Automated feeder design system}
For completeness, we define the trap transition matrix and summarize the automated feeder design system from \cite{mathiesen2017automatic}.

A trap can be modeled as a stochastic linear transition.
With $N$ stable poses and an additional special entry denoting discarded poses, the trap transition matrix, $T \in \R^{N+1 \times N+1}$, defines the probability that stable pose $s_j$ will transition to $s_i$ in the given trap, $T_{i,j} = P(s_i|s_j)$.
It follows that a column sums to one $\sum_i T_{i, j} = 1$, and that discarded poses remain discarded, $T_{N+1, N+1} = 1$.
Given a prior distribution, $P(s)$, over stable poses, where the last prior $P(s_{N+1})$ is zero, since no items are discarded initially, the pose distribution after the trap is $T P(s)$, and the transition matrix of a sequence of $K$ traps is simply found by chaining them, $T = T^1 T^2 \cdots T^K$.

Given a predefined set of $M$ traps represented by their transition matrices, the task for the automated feeder design system is to find a sequence of traps that results in a desired output pose distribution.
If all combinations of traps are allowed and explored in a brute-force manner, the search needs to explore $M^K$ combinations.
This is feasible when $K$ is \textit{very} low (2 or 3 trap slots)
and $M$ is in the range of thousands.



\subsection{Vision trap transition matrix}
In \cite{mathiesen2018optimisation}, dynamic simulation is used to estimate the transition matrix of mechanical traps. 
In this work, we aim to estimate transition matrices for a vision trap based on synthetic images, such that it can be incorporated seamlessly into the feeder design framework.
The vision trap is a rejection type trap, where parts either remain in the same stable pose or are discarded, that is $T_{i, i} + T_{N+1, i} = 1$, and the other entries are zero.

A decision needs to be made, whether or not to trigger the rejection mechanism based on the classifier's estimated stable pose distribution.
Based on a subset of the $N$ stable poses, $S_+ \subset S$, that should be allowed to pass through the trap, 
we define the condition as $P(S_+|I) > \tau$, 
where $\tau$ is a pre-defined model certainty threshold, 
and $P(S_+|I)$ is the sum of probabilities over the allowed set of stable poses $\sum_{s_i\in S_+} P(s_i | I)$ as estimated by the classifier.
The corresponding transition matrix is then estimated by
\begin{equation}
    T_{i,j} = 
    \begin{cases}
        \dfrac{1}{|X_j|} \sum_{I \in X_j} \delta_{P(S_+|I) > \tau}, & \text{if } j=i \leq N\\
        1-T_{j,j}, & \text{if } j\neq i=N+1\\
        1, & \text{if } j=i=N+1\\
        0, & \text{otherwise,}
    \end{cases}
\end{equation}
where $X_j$ is the set of synthetic image crops of stable pose $s_j$,
and $\delta_{[\text{condition}]}$ is $1$ if the condition holds and $0$ otherwise.
We use $\tau=0.9$ in our experiments.

Since we aim towards a fully autonomous configuration, and the interplay between traps is non-trivial, we relieve the user of deciding on $S_+$, by providing the feeder design system with all the possible, $2^N$, configurations.

\subsection{Stable pose reduction based on visual ambiguity}
\label{sec:pose-reduction}
As mentioned, the automated feeder design system needs to explore $M^K$ different configurations, where $M$ is the number of available traps, and $K$ is the maximum number of trap slots in the physical system.

Assuming there is one and only one vision trap, we would need to explore
$K M_v M_m^{K-1}$ different combinations, where 
$M_m$ is the number of mechanical traps (and their configurations),
$M_v = 2^N$ is the number of possible configurations of the vision trap,
and 
$K$ is because the vision trap can be placed in any of the trap slots.

Replacing a mechanical trap with a vision trap thus results in the same amount of configurations to explore, when $K M_v = M_m$, 
however, $M_v$ scales exponentially with $N$, which becomes prohibitive for large $N$.

To handle parts with many stable poses, sets of stable poses can be collapsed to reduce $N$.
Intuitively, we want to reduce $N$ while retaining as many meaningful configurations as possible.
Note that in the case of visual ambiguity between two stable poses, it would make sense to treat them as one by either having both or none of them in $S_+$, since having one and only one of them in $S_+$ would always lead to a low confidence, $P(S_+|I)$.
To avoid manual input to the pose reduction, we thus propose to reduce $N$ to a manageable number automatically based on visual ambiguity.

We obtain a probability confusion matrix, $C$, by running inference on the synthetic image crops
\begin{equation}
\label{eq:confusion_matrix}
    C_{i,j} = \dfrac{1}{|X_i|} \sum_{I\in X_i} P(s_j|I).
\end{equation}
We call $C$ a probability confusion matrix, since it is not an average over decisions, but rather an average over pose distributions.
For brevity, we will simply refer to $C$ as a confusion matrix from now on.


We then define a score for a confusion matrix as the log-likelihood of the true stable pose
\begin{equation}
    s(C) = \sum_{i=1}^N P(s_i) \log C_{i,i},
\end{equation}
and iteratively collapse the stable pose pair that maximizes the score, until $N$ has decreased to a desired size.
The score ensures that stable poses with high visual ambiguity will be collapsed before collapsing stable poses with low visual ambiguity.

An example of an initial confusion matrix and a reduced version is shown in Fig.~\ref{fig:confusion}.
Note that while in the example, $N$ is reduced to 4, the feeder designer can handle $2^N$ in the range of thousands, so a practical target for $N$ could be around 10 to 16.
Also note that some objects have many stable poses. A cube for instance has $6 \cdot 4 = 24$ stable poses, and then the reduction is necessary.
\section{Evaluation}
\label{sec:evaluation}

\begin{figure}
    \centering
    \includegraphics[width=0.7\linewidth, trim={0, 50, 0, 100}, clip]{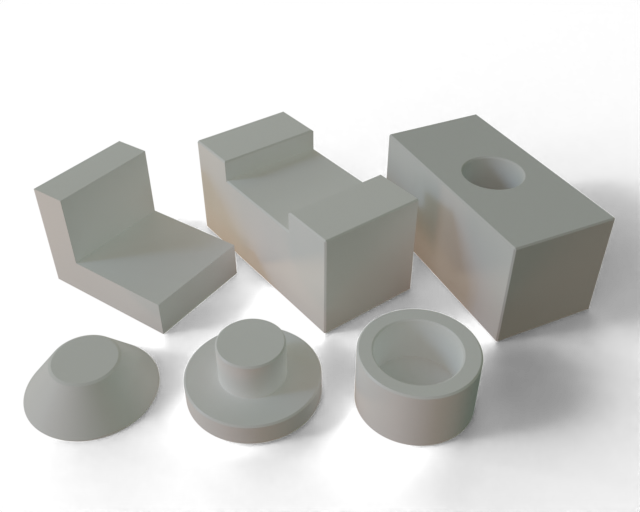}
    \caption{The part used for the evaluation from top left corner to bottom right are denoted: L-shape (LS), Rectangle w. indentation (RI), Rectangle w. hole (RH), truncated cone (TC), Top hat (TH) and Cap (CP).}
    \label{fig:objs}
\end{figure}


We evaluate the vision trap on a selection of six canonical parts which were originally published with~\cite{Boothroyd2005} as relevant parts for rejection type traps. The parts are shown in Fig.~\ref{fig:objs}. 
3D-models for the parts were drawn to be used for dynamic simulation as well as 3D-printing of the parts on a consumer-grade FDM printer.
The physical part properties used for the dynamic simulation were determined experimentally for the material type as in~\cite{mathiesen2018optimisation}. As described in Sec.~\ref{sec:method}, we use the dynamic simulation to identify stable poses and train vision models from synthetic images.

We then perform two experiments on the experimental setup shown in Fig.~\ref{fig:visiontrap}.
It is on the basis of visual ambiguities estimated on synthetic data that the feeder designer should configure the vision trap together with mechanical traps,
and thus our first experiment is a qualitative comparison between the confusion matrix on synthetic data and real data, providing an indication of the visual reality gap.
To obtain a confusion matrix on real data, we annotate the correct stable pose for the \textit{Cap} part in 20 real images with 141 instances in total and feed the images to the stable pose classifier. 
The confusion matrix is then calculated as in (\ref{eq:confusion_matrix}).

Our second and main experiment is a quantitative performance evaluation.
To evaluate the vision trap in isolation from mechanical traps, for each of the six canonical parts, we choose a stable pose with high confidence based on the estimated confusion matrix.
We pour approximately ten instances at a time from a box onto the beginning of the feeder, and let the instances vibrate into the vision trap until all instances have either been rejected or allowed through. The performance is examined for at least 100 instances per part.
The experiment is recorded and afterwards we manually note the number of instances that were correctly allowed to pass through (true positives), that were correctly discarded (true negatives), as well as instances that were incorrectly allowed through (false positives), and instances that were incorrectly discarded (false negatives).


\section{Results}
\label{sec:results}
\begin{figure*}
    \centering%
    \subfloat[\hspace{-0.8cm} ]{\label{fig:confReal}\includegraphics[width=.33\textwidth]{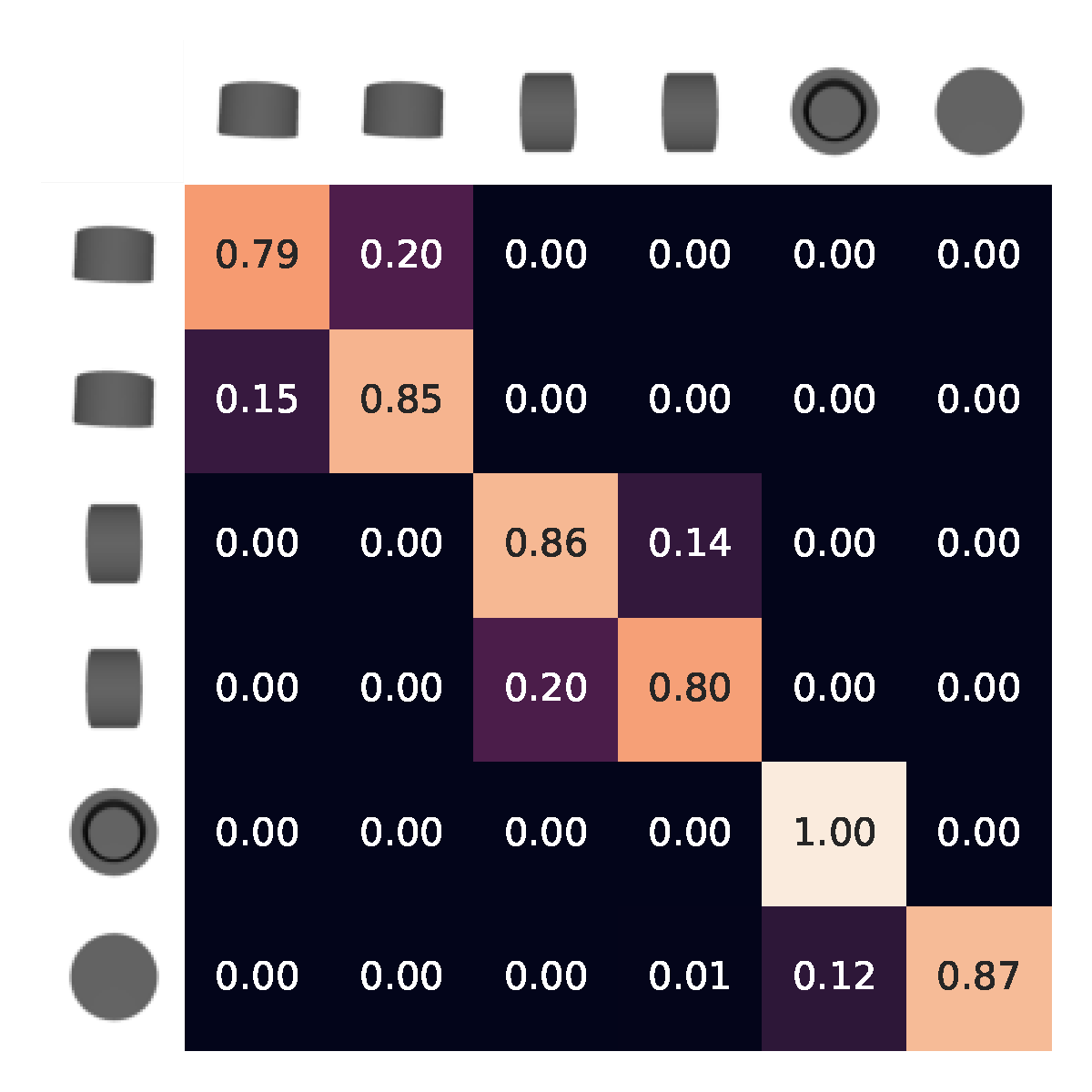}}\hspace{0.01\textwidth} \vrule height 5.7cm \hspace{0.1cm}%
    \subfloat[\hspace{-0.8cm} ]{\label{fig:confSynth}\includegraphics[width=.33\textwidth]{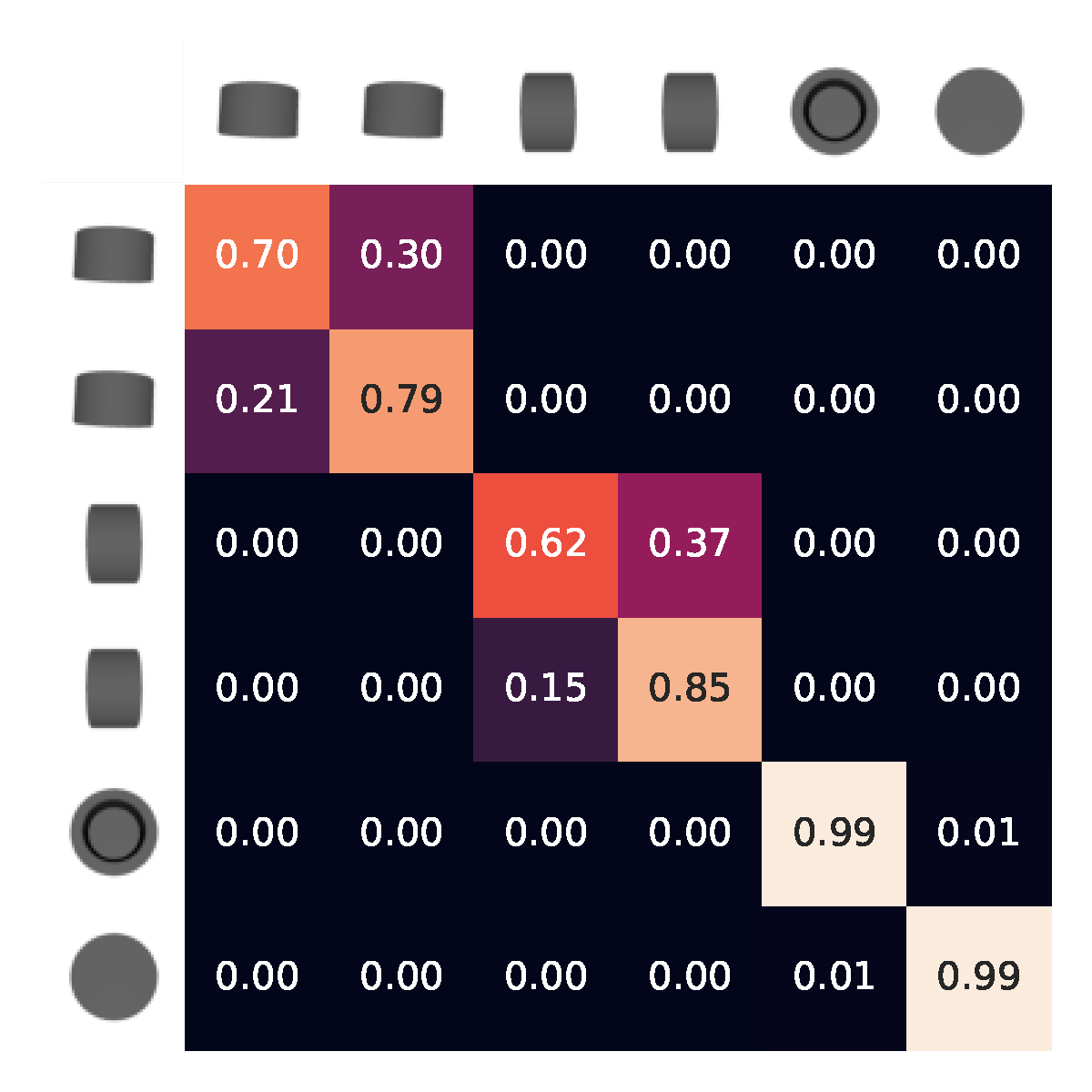}}%
    \subfloat[\hspace{-1.6cm} ]{\label{fig:confReduced}\includegraphics[width=.285\textwidth]{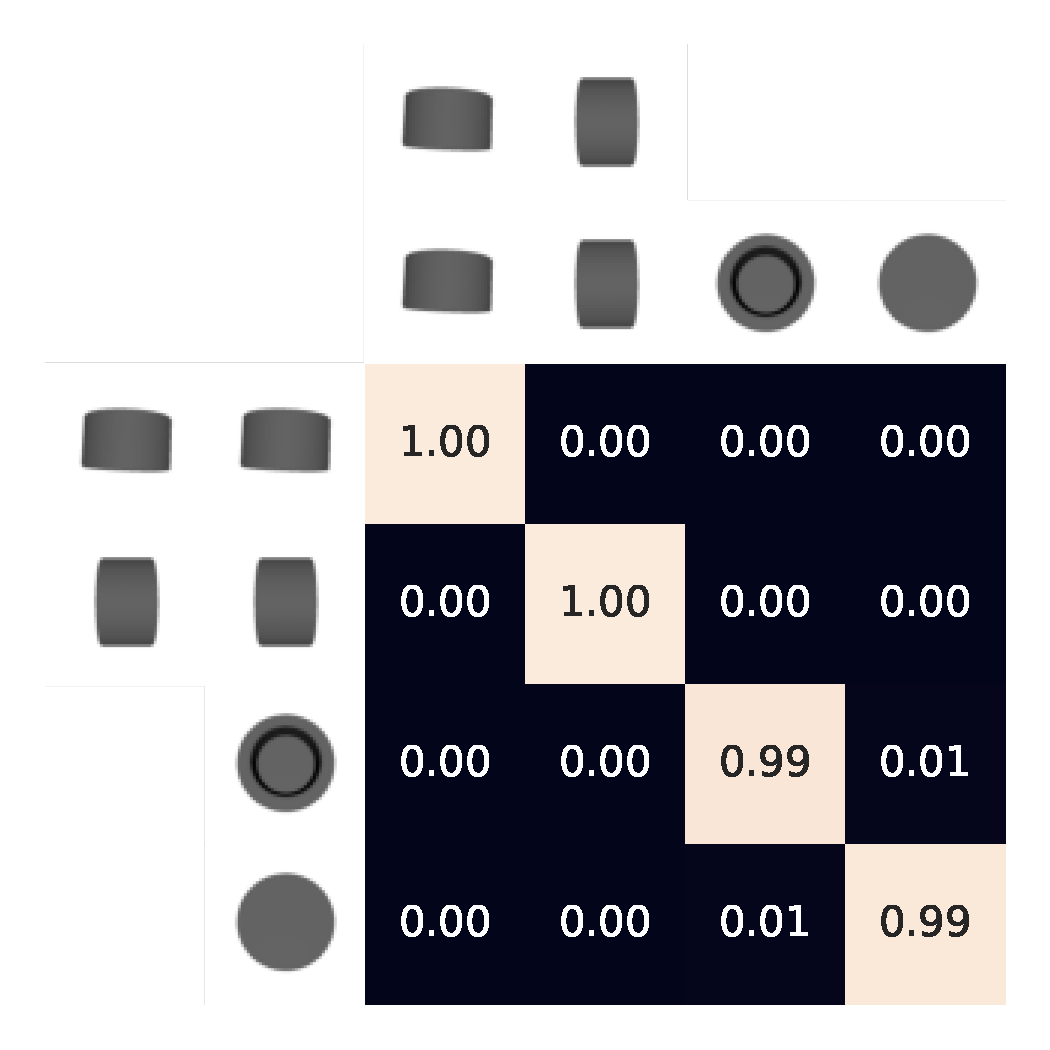}}%
    \caption{
        Confusion matrices from the model trained on synthetic images of part CP.
        \protect\subref{fig:confReal} is obtained from inference on manually annotated real images, which is not normally available and is only shown to validate
        \protect\subref{fig:confSynth} and \protect\subref{fig:confReduced}, which are obtained from inference on the synthetic images. 
        \protect\subref{fig:confReduced} is reduced to $N=4$ with the method described in Sec.~\ref{sec:pose-reduction}, automatically collapsing two pairs of stable poses based on visual ambiguity.
    }%
    \label{fig:confusion}
\end{figure*}

We present the results of our two experiments.
First, Fig.~\ref{fig:confusion} shows the confusion matrices for the \textit{Cap} part.
Second, Table~\ref{tab:trapEval} shows the performance of the vision trap under operation for each of the six parts. 
Finally, we discuss factors affecting how this approach can be applied in practice.   

\begin{table}
    \centering
    \caption{The evaluation of the vision trap for the six parts.}
    \begin{tabular}{l|rrrrrr}
    \hline
    Part name & CP & TH & TC & RI & RH & LS\\
    \hline
     True positives &    40 &       46 &          57 &             8 &            32 &        6 \\
     True negatives  &    63 &       62 &          41 &            93 &            65 &      108 \\
     False positives &     1 &        0 &           0 &             0 &             0 &        0 \\
     False negatives &     1 &        0 &          10 &             9 &            13 &        1 \\ \hline
     Total           &   105 &      108 &         108 &           110 &           110 &      115 \\
    \hline
    \end{tabular}
    
    \label{tab:trapEval}
\end{table}

The confusion matrix in Fig.~\ref{fig:confSynth} provides insight to the predicted capabilities of the trained classifier.
As described in Sec.~\ref{sec:pose-reduction}, the matrix shows the average probability the classifier contributes to the pose shown in the column, when presented with the pose shown in the row.
For stable pose visualization, we represent the stable pose, $s_i$, by the discrete pose in the connected component with the largest prior. 
The visualizations of the stable poses, as seen from above, are shown next to the rows and columns. 
From this matrix, it can be seen that there is significant ambiguity between the pose pairs 1 and 2 as well as 3 and 4. 
However, there is a high confidence for the individual poses 5 and 6, making these poses ideal candidates for desired stable poses

The confusion matrix found on real images, Fig.~\ref{fig:confReal}, is overall similar to the confusion matrix predicted on synthetic images, Fig.~\ref{fig:confSynth}, however, there is some difference introduced from the gap between synthetic and real images.
Perhaps most notably, the learnt model has more probability allocated to pose 5 when presented with pose 6 on the real data rather than the synthetic data. 
This may be related to circular patterns introduced by the 3D-printing method which to some degree match the features of pose 5, but are not modeled in the synthetic images. 
Note, however, that the confusion matrices do not need to match perfectly for the vision trap to be robust, since the target confidence $\tau$ has to be reached to allow an instance to pass through the trap.

Fig.~\ref{fig:confReduced} shows how the number of stable poses can be reduced to make the following optimization problem in the feeder design system feasible, while retaining as much information as possible.
As described in Sec.~\ref{sec:pose-reduction}, a feasible number of stable poses for the feeder design system is probably around 10-16, but the reduction is shown here for a smaller number of poses for clarity.

\begin{figure*}
    \centering%
    \hfill%
    \subfloat[]{\label{fig:collateral}\includegraphics[width=.315\textwidth, trim={110, 170, 110, 120}, clip]{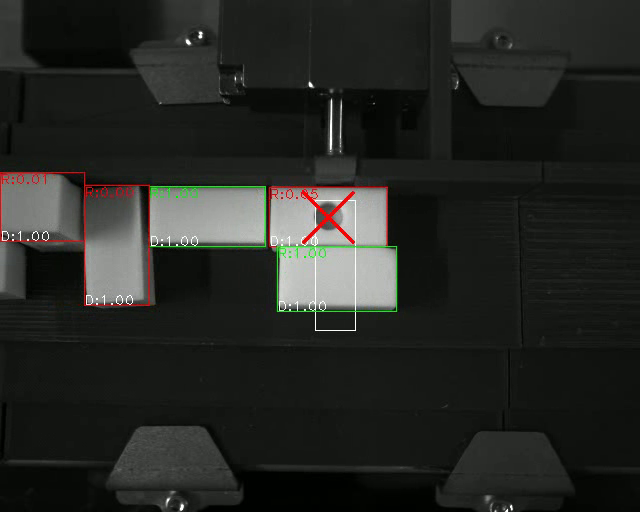}}\hfill%
    \subfloat[]{\label{fig:badclass}\includegraphics[width=.315\textwidth, trim={110, 170, 110, 120}, clip]{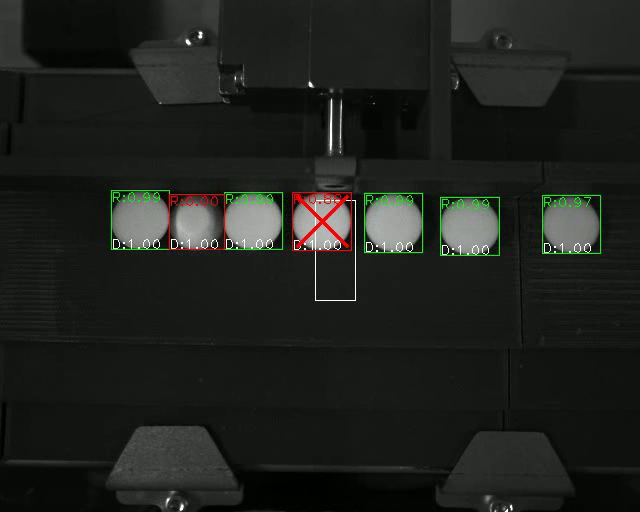}}\hfill%
    \subfloat[]{\label{fig:unstable}\includegraphics[width=.315\textwidth, trim={110, 170, 110, 120}, clip]{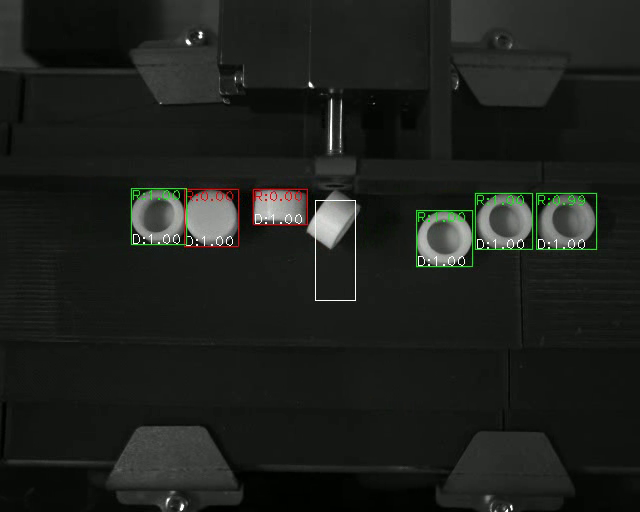}}\hfill%
    \caption{
        The three types of observed failure cases.
        \protect\subref{fig:collateral} shows the most common type, where the correct rejection of a negative results in rejecting a positive as well, which then counts as a false negative in the results.
        \protect\subref{fig:badclass} shows an example, where the certainty of the stable pose classifier falls below $\tau$, such that the instance is wrongfully discarded (false negative).
        \protect\subref{fig:unstable} shows the only false positive during the experiments. In this case, the pose is temporarily unstable after the rejection of a negative. The unstable pose is not represented by any stable pose and is thus not seen by the detector during training.
    }
    \label{fig:failure_cases}
\end{figure*}

The results of our performance test on the six canonical parts are shown in Table~\ref{tab:trapEval}. 
The results show that a notable number of positives are rejected (false negatives).
We observe two types of false negatives. 
The most common is when rejection of a correctly identified negative leads to simultaneously rejecting a true positive, as seen in Fig.~\ref{fig:collateral}, and account for 29 out of the 34 false negatives.
The remaining five false positives arise from the model simply not being certain enough about the classification to allow the instance to pass through, as seen in Fig.~\ref{fig:badclass}.
False negatives will of course impact the throughput of the system, but are much less critical than false positives, which may require manual intervention to reset the system.
Among more than 600 instances that are fed through the vision trap, there is only one false positive which is shown in Fig.~\ref{fig:unstable}.
This singular case arises because the rejection of a negative temporarily causes an unstable pose, which the detector has not seen during simulation and therefore misses. 
An approach to reduce false positives further could thus be to include unstable poses to the training data for the detector and add a custom unstable class in the classifier.

The success of the vision trap relies on the accuracy of the various models used for simulation and training, including 3D models, material properties, etc., but as the manufacturing industry becomes increasingly digitized it becomes more reasonable to assume that accurate models are available. 
For a single part,
the dynamic simulation to obtain stable poses was run in a few hours,
and the vision detector and classifier were subsequently trained in around an hour on a single GPU. 
In the context of the broader feeder design framework, this enables very fast identification of which stable poses, if any, the vision trap can feed by itself, and thus whether mechanical traps are even needed for a specific part.

Note that, in contrast to mechanical traps, which need to be designed and constructed or at least adjusted and then physically installed, the vision trap is configured purely in software, reducing the setup time and cost, which could make the vision trap preferable even if the desired behaviour could be solved by a set of mechanical traps.

More importantly, it is very flexible compared to mechanical traps,
as it automatically provides a set of different configurations with \textit{varied behaviors}.
This is in stark contrast to mechanical traps, which are designed and optimized towards specific behaviours.
A single vision trap can also potentially deliver the parts in a wider range of desirable stable poses than a sequence of dedicated mechanical traps from~\cite{Boothroyd2005}.
For example, \cite{Boothroyd2005} has one trap to reject pose 1-4 in Fig.~\ref{fig:confSynth}, and another trap to reject pose 6, but no trap to reject pose 5.
Based on Fig.~\ref{fig:confReduced}, the vision trap is able to provide either 1-2, 3-4, 5 or 6 with a single trap.
The capacity of the vision model could be further improved by providing multiple views of the feeding section, either with multiple cameras or with a mirror.

In many cases, the vision trap alone could be sufficient.
However, when that is not the case, the vision trap might still reduce the design problem significantly for finding a mechanical trap solution.
The confusion matrix could also potentially be used to limit the search of relevant mechanical trap types.
Another potential use could be providing part designers a tool, or even an automatic unit test, for determining whether a new part can easily be fed or whether it would be worth redesigning the parts with more descriptive features.

The feeding rate during the experiments is approximately one part per second but can be improved by optimizing the vibration angle and frequency.
The vision pipeline runs at around 13Hz on an RTX 2070 GPU, 
but the computational requirements could be greatly reduced by 
sharing the backbone between the detection and classification model,
using a smaller backbone,
and reducing the size of the input image by using a tighter crop around the rejection region.
Note that the images in Fig.~\ref{fig:failure_cases} are cropped versions of the input image.

\section{Conclusion}
\label{sec:conclusion}
This work proposed a novel approach to the use of vision for enhancing the flexibility of vibratory feeders. 
A vision trap was presented which can be configured from dynamic simulation and synthetic images. 
The core strength of the approach is predicting the capabilities to distinguish between stable poses in different configurations, enabling an automated feeder design framework to jointly configure and combine the vision trap with suiting mechanical traps to obtain a desired trap as a whole.
The vision trap was evaluated on six canonical parts with promising results. 
The primary type of observed errors were false negatives, which are not critical in small numbers and could mostly be attributed to mechanical considerations related to the actuated rejection mechanism. 
Only one false positive occurred among more than 600 instances and we have provided a concrete approach to address the root cause. 
The approach thus shows great promise for fully autonomous feeding configuration, a significant step towards economically feasible low volume high variance productions.

{\small
\bibliographystyle{IEEEtran}
\bibliography{references}

\begin{thebibliography}{10}
\providecommand{\url}[1]{#1}
\csname url@rmstyle\endcsname
\providecommand{\newblock}{\relax}
\providecommand{\bibinfo}[2]{#2}
\providecommand\BIBentrySTDinterwordspacing{\spaceskip=0pt\relax}
\providecommand\BIBentryALTinterwordstretchfactor{4}
\providecommand\BIBentryALTinterwordspacing{\spaceskip=\fontdimen2\font plus
\BIBentryALTinterwordstretchfactor\fontdimen3\font minus
  \fontdimen4\font\relax}
\providecommand\BIBforeignlanguage[2]{{%
\expandafter\ifx\csname l@#1\endcsname\relax
\typeout{** WARNING: IEEEtran.bst: No hyphenation pattern has been}%
\typeout{** loaded for the language `#1'. Using the pattern for}%
\typeout{** the default language instead.}%
\else
\language=\csname l@#1\endcsname
\fi
#2}}

\bibitem{hodavn2020bop}
T.~Hoda{\v{n}}, M.~Sundermeyer, B.~Drost, Y.~Labb{\'e}, E.~Brachmann,
  F.~Michel, C.~Rother, and J.~Matas, ``Bop challenge 2020 on 6d object
  localization,'' in \emph{European Conference on Computer Vision}.\hskip 1em
  plus 0.5em minus 0.4em\relax Springer, 2020, pp. 577--594.

\bibitem{hagelskjaer2018does}
F.~Hagelskj{\ae}r, A.~G. Buch, and N.~Kr{\"u}ger, ``Does vision work well
  enough for industry?'' in \emph{VISIGRAPP (4: VISAPP)}, 2018, pp. 198--205.

\bibitem{mathiesen2018optimisation}
S.~Mathiesen, L.~C. S{\o}rensen, D.~Kraft, and L.-P. Ellekilde, ``Optimisation
  of trap design for vibratory bowl feeders,'' in \emph{2018 IEEE international
  conference on robotics and automation (ICRA)}.\hskip 1em plus 0.5em minus
  0.4em\relax IEEE, 2018, pp. 3467--3474.

\bibitem{mathiesen2017automatic}
S.~Mathiesen and L.-P. Ellekilde, ``Automatic selection and sequencing of traps
  for vibratory feeders,'' in \emph{7th International Conference on Simulation
  and Modeling Methodologies Technologies and Applications (SIMULTECH)}.\hskip
  1em plus 0.5em minus 0.4em\relax SCITEPRESS Digital Library, 2017, pp.
  145--154.

\bibitem{Boothroyd2005}
G.~Boothroyd, \emph{Assembly Automation and Product design}, 2nd~ed.\hskip 1em
  plus 0.5em minus 0.4em\relax CRC Press, 2005.

\bibitem{malik2019advances}
A.~A. Malik, M.~V. Andersen, and A.~Bilberg, ``Advances in machine vision for
  flexible feeding of assembly parts,'' \emph{Procedia Manufacturing}, vol.~38,
  pp. 1228--1235, 2019.

\bibitem{hagelskjaer2019using}
F.~Hagelskj{\ae}r, T.~R. Savarimuthu, N.~Kr{\"u}ger, and A.~G. Buch, ``Using
  spatial constraints for fast set-up of precise pose estimation in an
  industrial setting,'' in \emph{2019 IEEE 15th International Conference on
  Automation Science and Engineering (CASE)}.\hskip 1em plus 0.5em minus
  0.4em\relax IEEE, 2019, pp. 1308--1314.

\bibitem{okorn2020learning}
B.~Okorn, M.~Xu, M.~Hebert, and D.~Held, ``Learning orientation distributions
  for object pose estimation,'' in \emph{2020 IEEE/RSJ International Conference
  on Intelligent Robots and Systems (IROS)}.\hskip 1em plus 0.5em minus
  0.4em\relax IEEE, 2020, pp. 10\,580--10\,587.

\bibitem{murch1971predicting}
L.~Murch and G.~Boothroyd, ``Predicting efficiency of parts orienting
  systems,'' \emph{Automation}, vol.~18, pp. 55--57, 1971.

\bibitem{christiansen1996automated}
A.~D. Christiansen, A.~D. Edwards, and C.~A.~C. Coello, ``Automated design of
  part feeders using a genetic algorithm,'' in \emph{Proceedings, International
  Conference on Robotics and Automation}, vol.~1.\hskip 1em plus 0.5em minus
  0.4em\relax IEEE, 1996, pp. 846--851.

\bibitem{edmondson2001flexible}
N.~Edmondson and A.~Redford, ``Flexible parts feeding for flexible assembly,''
  \emph{International Journal of Production Research}, vol.~39, no.~11, pp.
  2279--2294, 2001.

\bibitem{loy2010new}
M.~Loy and G.~Reinhart, ``A new modular feeding system and its economic scope
  of application,'' \emph{Production engineering}, vol.~4, no.~4, pp. 357--362,
  2010.

\bibitem{kolditz2021flexible}
T.~Kolditz, P.~M{\"u}ller, D.~Bansmann, and A.~Raatz, ``Flexible aerodynamic
  part feeding using high-speed image processing,'' in \emph{Congress of the
  German Academic Association for Production Technology}.\hskip 1em plus 0.5em
  minus 0.4em\relax Springer, 2021, pp. 403--411.

\bibitem{posecnn}
Y.~Xiang, T.~Schmidt, V.~Narayanan, and D.~Fox, ``Posecnn: A convolutional
  neural network for 6d object pose estimation in cluttered scenes,'' 2018.

\bibitem{surfemb}
R.~L. Haugaard and A.~G. Buch, ``Surfemb: Dense and continuous correspondence
  distributions for object pose estimation with learnt surface embeddings,'' in
  \emph{Proceedings of the IEEE/CVF Conference on Computer Vision and Pattern
  Recognition}, 2022, pp. 6749--6758.

\bibitem{sundermeyer}
M.~Sundermeyer, Z.-C. Marton, M.~Durner, M.~Brucker, and R.~Triebel, ``Implicit
  3d orientation learning for 6d object detection from rgb images,'' in
  \emph{Proceedings of the european conference on computer vision (ECCV)},
  2018, pp. 699--715.

\bibitem{faster-rcnn}
S.~Ren, K.~He, R.~Girshick, and J.~Sun, ``Faster r-cnn: Towards real-time
  object detection with region proposal networks,'' \emph{Advances in neural
  information processing systems}, vol.~28, 2015.

\bibitem{resnet}
K.~He, X.~Zhang, S.~Ren, and J.~Sun, ``Deep residual learning for image
  recognition,'' in \emph{Proceedings of the IEEE conference on computer vision
  and pattern recognition}, 2016, pp. 770--778.

\bibitem{imagenet}
J.~Deng, W.~Dong, R.~Socher, L.-J. Li, K.~Li, and L.~Fei-Fei, ``Imagenet: A
  large-scale hierarchical image database,'' in \emph{2009 IEEE conference on
  computer vision and pattern recognition}.\hskip 1em plus 0.5em minus
  0.4em\relax Ieee, 2009, pp. 248--255.

\bibitem{coco-dataset}
T.-Y. Lin, M.~Maire, S.~Belongie, J.~Hays, P.~Perona, D.~Ramanan,
  P.~Doll{\'a}r, and C.~L. Zitnick, ``Microsoft coco: Common objects in
  context,'' in \emph{European conference on computer vision}.\hskip 1em plus
  0.5em minus 0.4em\relax Springer, 2014, pp. 740--755.

\bibitem{adam}
D.~P. Kingma and J.~Ba, ``Adam: A method for stochastic optimization,'' in
  \emph{ICLR (Poster)}, 2015.

\end{thebibliography}
}

\end{document}